\documentclass[conference]{IEEEtran}
\IEEEoverridecommandlockouts

\usepackage[table]{xcolor}
\usepackage{amsmath}

\usepackage{lscape}
\usepackage[T1]{fontenc} 
\usepackage{booktabs}
\usepackage[ruled,vlined]{algorithm2e}
\usepackage{wrapfig}
\usepackage{csquotes}
\usepackage{float} 
\usepackage{hyperref}
\hypersetup{
    colorlinks=true,
    linkcolor=cyan,
    filecolor=black,      
    urlcolor=cyan,
    citecolor=cyan,
}

\usepackage{cite}
\usepackage{amsmath,amssymb,amsfonts}
\usepackage{graphicx}
\usepackage{textcomp}
\usepackage{xcolor}
\begin{document}

\title{Understanding YouTube Communities via Subscription-based Channel Embeddings
}

\author{\IEEEauthorblockN{Sam Clark}
\IEEEauthorblockA{ 
\textit{Independent Researcher}\\
Seattle, United States \\
sclark.uw@gmail.com \\
}
\and
\IEEEauthorblockN{Anna Zaitsev}
\IEEEauthorblockA{\textit{The School of Information} \\
\textit{University of California, Berkeley}\\
Berkeley, United States \\
anna.zaitsev@ischool.berkeley.edu}
}

\maketitle

\begin{abstract}
YouTube is an important source of news and entertainment worldwide, but the scale makes it challenging to study the ideas and topics being discussed on the platform. This paper presents new methods to discover and classify YouTube channels which enable the analysis of communities and categories on the platform using orders of magnitude more channels than have been used in previous studies. Instead of using channel and video data as features for classification as other researchers have, these methods use a self-supervised learning approach that leverages the public subscription pages of commenters. We test the classification method on the task of predicting the political lean of YouTube news channels and find that it outperforms the previous best model on the task. Further experiments also show that there are important advantages to using commenter subscriptions to discover channels. The subscription data, along with an iterative approach, is applied to discover, to our current understanding, the most comprehensive set of English language socio-political YouTube channels yet to be analyzed. We experiment with predicting more fine grained political tags for channels using a previously annotated dataset and find that our model performs better than the average individual human reviewer for most of the top tags. This fine grained political tag model is then applied to the newly discovered English language socio-political channels to create a new dataset to analyze the amount of traffic going to different political content. The data shows that some tags, such as "Partisan Right" and "Conspiracy", are significantly under represented when looking only at the most popular socio-political channels. Through the use of our methods, we are able to get a much more accurate picture of the size of these communities on YouTube.
\end{abstract}


\section{Introduction}
YouTube, with over 2 billion monthly active users \cite{youtube2020youtube}, has become one of the largest social media platforms and the largest video sharing platform in the world. The platform was created as a video sharing resource for independent content creators. These creators have made the platform popular, producing a very diverse set of content from political commentary to makeup tutorials, music videos, vlogs, how-to-videos, and gaming. Nowadays nearly every topic and category imaginable is covered on the platform, including content from traditional TV and network media outlets. Users have embraced YouTube as their source of entertainment as well as news. According to a recent study by Pew Research Center, 26 percent of adults in the United States get news from YouTube \cite{stocking2020share}.

As YouTube has grown to become a source for entertainment and news around the world, the scale has also presented challenges for studying the ideas being shared and communities being formed on the platform. On cable and network TV, an individual can be aware of all the channels available in a region and have a sense of the type of content common on them. In addition, Nielsen\footnote{\hyperlink{https://www.nielsen.com/us/en/}{https://www.nielsen.com/us/en/}}, an information and data company, tracks TV ratings, so this data can be used to compare the audience size for these different types of content. With millions of channels on YouTube, a similar level of knowledge collected manually by an individual (or even group of researchers) is impossible.

Prior studies have taken two different approaches: either focus on breadth or depth of YouTube data. Studies that focus on the breadth of YouTube attempt to analyze the content, for example, political channels, by focusing on a set of the most popular channels \cite{stocking2020share, dinkov2019predicting}. This limitation means that a significant number of smaller channels, which in aggregate could have more traffic than the popular channels, are excluded from the analysis. Conversely, the papers that focus on the depth of individual communities and attempt to understand a discreet subset of YouTube content, provide a detailed insight into various niche communities \cite{ribeiro2019auditing, papadamou2020understanding}. However, these studies offer no comparative data which would provide information on the proportionality of these niche communities. Additionally, researchers using both of these approaches have failed to measure the overall coverage of community or category content their studies account for. 

This study proposes novel methods for automated channel discovery and classification, which combined enable the ability to analyze orders of magnitude more channels when studying communities or categories on YouTube. Both methods leverage "commenter subscriptions", that is the subscriptions displayed on the public profile pages of commenters. This data is used as follows:
\begin{enumerate}
    \item The average commenter subscribes to 200 channels, so this data can be used to identify new channels quickly.
    \item It is also the input data for a self-supervised machine learning approach that generates embeddings for each channel. These embeddings are then used along with labeled channel data to classify characteristics of unlabeled channels. 
\end{enumerate}

We believe that this approach is effective for studying a variety of categories and communities. However, due to the increasing importance of YouTube as a source of news \cite{stocking2020share} and the number of datasets focusing on political YouTube, we focus our experiments on YouTube channels that discuss political and cultural issues.

The efficacy of our methods are tested through the following experiments:
\begin{enumerate}
    \item We use our method for discovering new channels to identify channels that fit the criteria for an existing socio-political dataset \cite{ledwich2020algorithmic}. This results in order of magnitude more channels being discovered than were in the original dataset. We then attempt to measure what percentage of the overall socio-political landscape these new channels cover.
    \item We test our method of classifying channels on the task of classifying the granular political and ideological labels given to channels in the previous socio-political dataset \cite{ledwich2020algorithmic}. Three reviewers manually generated the labels, and we find that our model outperforms the average reviewer for the majority of labels.
    \item Finally, we compare our channel classification method to another predictive model \cite{dinkov2019predicting}. This model uses a variety of metadata features, video transcripts and is even able to achieve an increase in performance from audio signals. However, on a set of channels that both approaches can make predictions on, our method achieves an accuracy of 83.8\% compared to 73.0\% by theirs.
\end{enumerate}

Along with testing the efficacy of our methods, we also do a short analysis of head vs. tail socio-political channels to motivate the importance of doing more comprehensive studies of communities on YouTube. We do this using a combination of channels from the prior study \cite{ledwich2020algorithmic} along with the newly discovered and classified socio-political channels. We find that just focusing on the most popular socio-political channels, those with over 500K subscribers, results in a very inaccurate view of the percentage of traffic going to 'Partisan Right' and 'Conspiracy' channels.

\section{Related Work}
Previous research on YouTube content has primarily focused on either classifying videos or channels. Large video datasets made available by YouTube \cite{karpathy2014large, abu2016youtube} have led to significant research on machine learning approaches to video classification \cite{karpathy2014large, Chen2017MultimodalAF, huang2018makes}. Other studies have used metadata and video comments for video classification tasks, such as identifying conspiracy videos \cite{faddoul2020longitudinal} and investigating 'incel' content \cite{papadamou2020understanding}.

Many channels upload videos that exclusively cover a small set of topics. Researchers have taken advantage of this by classifying content at the channel level instead of the video level. Manual classification of channels has been used to study right-wing content \cite{ribeiro2019auditing}, as well as the full spectrum of political content \cite{ledwich2020algorithmic, stocking2020share}. Machine learning approaches have been used as well, such as to predict the political lean of a channel \cite{dinkov2019predicting}, to predict whether a channel covers socio-political content along with their ideology \cite{laukemper2020classifying}, and by YouTube internally to predict broad categories \cite{simonet2013classifying}. However, the later uses a variety of data available only to the platform.

There have been a variety of studies that use natural language processing to classify political content outside of YouTube as well. For example, studies have experimented with predicting the political ideology of Twitter users \cite{preoctiuc2017beyond} and shown how one can identify biased, hyperpartisan news outlets, based on the writing style \cite{potthast2018stylometric}. Furthermore, datasets have been compiled to provide information on news outlets and their reliability, based on fact-checked news articles \cite{horne2018sampling} which have then been utilized to predict the bias and reliability of news outlets \cite{baly2018predicting, baly2019multi}.

\section{Channel Classification and Discovery Methods}
\subsection{Data Collection}
\label{datacollection}
\begin{figure}[ht]
\centering
    \includegraphics[width=\linewidth]{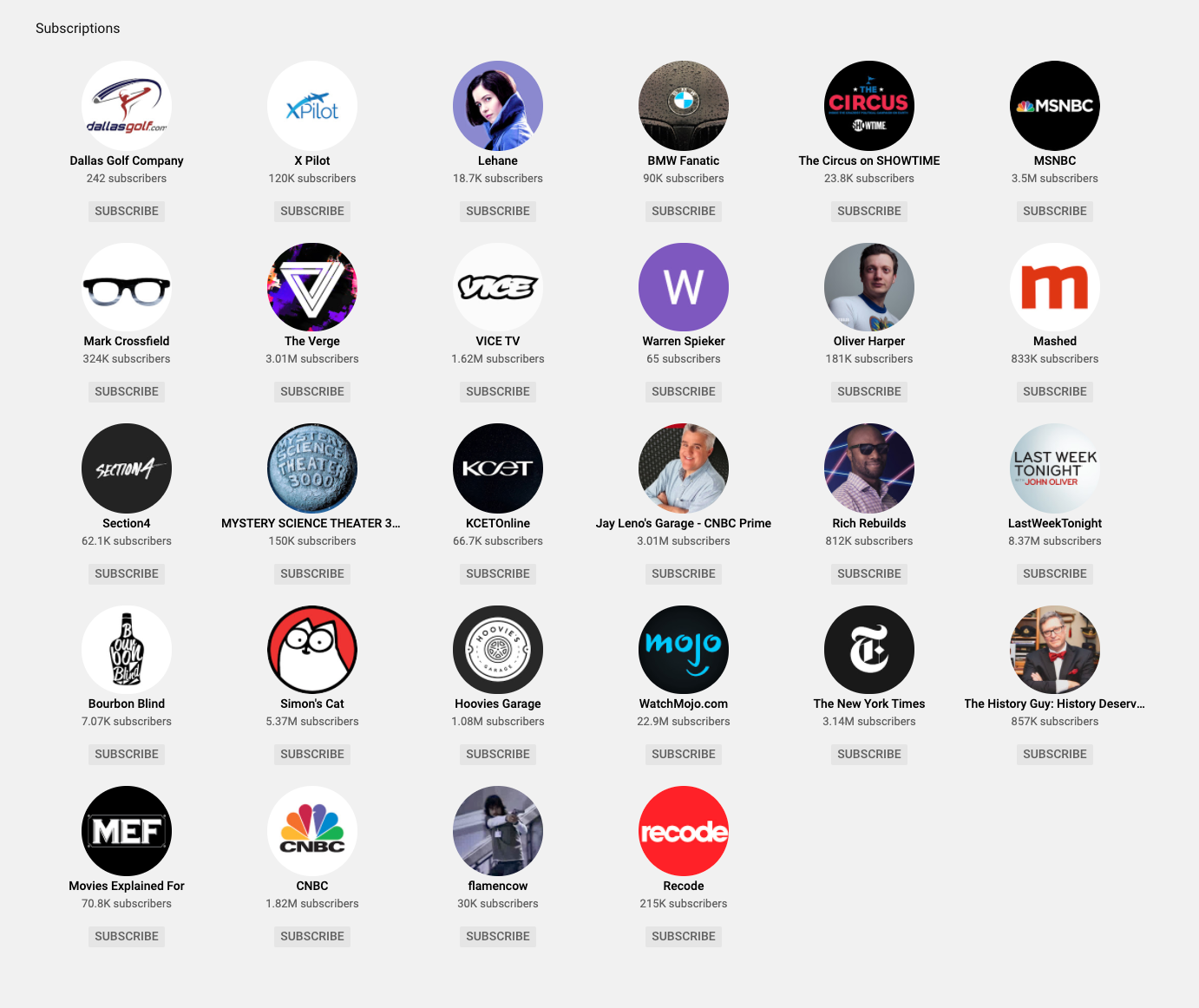}
    \caption{A subscription page}
    \label{subs}
\end{figure}

The critical piece of data our channel classification and discovery methods rely on is video "commenter subscriptions". These are the channels that commenters are subscribed to, and for a large percentage of commenters, they are visible to the public on their profile page. Figure \ref{subs} shows an example of a users subscription page. 

To collect this data for a channel, we do the following:
\begin{itemize}
    \item We query the thirty most recently uploaded videos for the channel.
    \item We query up to 100 comments for each of the ten most viewed of these videos.
    \item We query subscription data for all of these commenters that have public subscription pages (which we have found is roughly 30\% of commenters)
    \begin{enumerate}
        \item We use a method that only returns 30 subscriptions if they are available. We call these “sample commenter subscriptions” since they frequently do not cover all channels a commenter is subscribed to.
        \item However, for up to 10 of the commenters with subscriptions, we use a more computationally intensive method to query all available subscriptions.  On average, a user has 210 subscriptions, so this results in a significant increase in subscriptions. We call these “full commenter subscriptions”.
    \end{enumerate} 
\end{itemize}

\subsection{chan2vec Embeddings}

There are a variety of ways commenter subscriptions can be used for classification or topic modeling of channels. Our approach was inspired by work done at Instagram to create account embeddings for a similar purpose \cite{medvedev2019instagram}. We decided to infer channel embeddings from commenter subscriptions using a word2vec-like embedding framework we call \textit{chan2vec}. Typically, word2vec learns vectors (commonly called embeddings in this scenario) for words based on the context they appear across a corpus of sentences \cite{mikolov2013efficient}. The learning algorithm's objective is to learn vectors such that words that are semantically similar have similar vectors (specifically high cosine-similarity). For chan2vec, we treat the sequence of YouTube channel IDs subscribed to by a commenter as equivalent to a "sentence of words" and our "corpus of sentences" consists of commenters we have identified across a variety of channels and videos. 

In order to learn useful embeddings for a channel, we must have subscription information for enough of the subscribers of the channel to gain the proper signal. We set this threshold for the minimum commenter subscriptions to five, but further analysis is necessary to determine what the optimal threshold is.

To generate the embeddings used for chan2vec, we create "sentences" for each unique commenter we have collected subscription data for. We filter out all channels from these "sentences" that occur less than five times in the dataset, and we filter out all "sentences" that have less than three channels in them (meaning the commenter is subscribed to less than three channels). It is not clear how channels are ordered on public subscription pages, so we randomly shuffle the order of channels in "sentences" to eliminate any bias that might be introduced by the default ordering.

Unless otherwise specified, we use continuous-bag-of-words, a window size of 8, and 200 dimension vectors along with the other default word2vec parameters. Many of our experiments involve small datasets, so we do not have the ability to do a significant amount of parameter optimization without the risk of overfitting.

\subsection{Channel Classification}

In order to use chan2vec embeddings for classification, we must first gather commenter subscriptions for channels in a labeled dataset as well as the unlabeled channels we would like to classify. We do this using the method described in the previous section, section \ref{datacollection}. There is no guarantee that this will result in enough subscriptions collected for each of these channels in order to generate embeddings for them. 

However, in experiment \ref{dinkovcomparsion} we find that 67\% of channels with 1K+ subscribers and 87\% of channels with 10K+ subscribers can be found using this method. In that experiment we also introduce alternative methods to increase coverage.

After collecting the data, we then use chan2vec to generate embeddings. We then use these embeddings along with K-nearest-neighbors (KNN) and a dataset of labeled channels to make predictions. Specifically, let's say our labeled dataset consists of a set of channels each given a label of "1" if they have some descriptor D or "0" if they do not. If we want to predict whether channel C has descriptor D, we find the K closest channels in the labeled dataset to C using our chan2vec embeddings. Our prediction for C is then the average label of the K closest channels. For example, if we are trying to predict whether a channel is political, it is the percentage of the K nearest labeled channels that are labeled political.  We call the process of using KNN with chan2vec embeddings “chan2vec-knn”. As with word2vec, we use cosine-distance as our distance measure between any two channel embeddings. For most of our experiments, we use a K value of 5 or 10, but there is room for more experimentation to be done here. The threshold for the number of nearest neighbours that must have a positive label before making a positive prediction varies by experiment as well.

\subsection{Channel Discovery}

One challenge of analyzing communities or categories on YouTube is identifying channels that are members of them. YouTube does not provide granular categories or ways to browse categories extensively. Furthermore, even if the platform would provide more granular filtering features, only focusing on the larger channels gives limited coverage of all content given the scale of the platform.

In past studies of communities and categories on YouTube, two primary approaches have been used for identifying channels to focus on:
\begin{enumerate}
    \item Start with a set of seed channels, then manually review channels YouTube recommends as similar or that have videos frequently recommended for these seed channel videos. Add channels that are found to fit in the community or category being studied back into this set and repeat the process until saturation is reached or resources are exhausted, i.e., snowball sampling method \cite{lee2006statistical}.
    \item Create a set of keywords associated with the community or category under investigation, search YouTube for these keywords, then manually review the videos or channels or both, returned.
\end{enumerate}
  
These approaches have two major problems. First, these approaches require a manual review of channels. The manual review significantly limits the number of channels that can be found due to how time consuming it is. Second, YouTube's search and recommendation systems are less likely to return channels with a small number of subscribers and channels that post content that comes close to violating the platforms terms-of-service (and in some cases restrict them all together).

We propose a method for discovering channels that are members of specific categories or communities that does not suffer from these problems and makes it possible to analyze orders of magnitude more channels than have been analyzed in previous studies.

Our method for channel discovery is a simple extension of the chan2vec-knn and data collection methods described already. It takes advantage of the fact that for each new channel that commenter subscriptions are queried for, subscription data is gathered about many other channels that have co-occurring subscriptions with it. Due to the large number of subscriptions the average commenter has, new channels can be discovered at a rapid rate.

The method consists of two parts. It first uses an iterative approach that casts a wide net to identify "candidate" channels. That is, channels likely to fit in the category or community of interest. When this process finishes, a more accurate model is then used to predict which of the candidate channels are part of the category or community. These two algorithms are described next and presented in Figure \ref{algo1} and Figure \ref{algo2}. 

Formally, to discover channels with property P, we first create a set of channels that have been labeled as having property P (positive) or not having property P (negative). Let us call this set of labeled channels L. During our channel discovery process, we will keep track of a set of discovered channels that have been identified as being "candidates" for having property P. Let us call this set of candidate channels C. As an arbitrary step we will add all channels from L into C (even though we already know which are positive and negative).

An iterative approach (also described in Figure \ref{algo1}) is then used in which we first query commenter subscription data for all channels in C that have not previously had data queried for them. Let us call these candidate channels without data queried Ci and commenter subscriptions queried for them Si (where i = the number of the iteration). We then combine all commenter subscription data that has been previously queried to create S (so for the first iteration S = S1, but for later iterations S = S1 + S2 + .. Si).  The chan2vec process is then used to create embeddings E for all channels in S that have over a minimum number of commenter subscriptions. As mentioned earlier, this will result in embeddings for many more channels than are in C given the significant number of channels the average commenter is subscribed to.

For all channels with embeddings in E, KNN is then used with the set of labeled channels L to predict whether each channel has property P. Channels predicted to have property P are then added to the set of candidate channels C. If very few new channels were added to C, then the iterative process stops. Otherwise, another iteration is started.

Finally, after the iterative process finishes, a final set of predictions needs to be made. The algorithms is presented in Figure \ref{algo2}. We use the set of embeddings E generated from the last iteration and apply KNN to make predictions for all channels in C2..Cn, where n is the number of the final iteration (we exclude C1 since this contains all channels from L). Only those predicted to have property P in this final prediction are included in our final set D of discovered channels. 

\begin{algorithm}[h]
\SetAlgoLined
\(C_{1} \leftarrow C\) \\
\(i \leftarrow 1\) \\
\textbf{repeat}\\
\hskip2em \(S_{i} \leftarrow query\_commenter\_subs(C_{i})\)\\
\hskip2em \(S \leftarrow S_{1}\cup S_{2} \cup ...\cup S_{i}\)\\
\hskip2em \(E \leftarrow chan2vec(S)\)\\
\hskip2em \(C_{i+1} \leftarrow \emptyset\)\\
\hskip2em \(\textbf{for}\ x \in E \ \textbf{do}\) \\
\hskip4em \(\textbf{if}\ x\ channel \notin C\ \textbf{then}\) \\
\hskip6em \( y \leftarrow chan2vec\_knn(L, E, x)\)\\
\hskip6em \( \textbf{if}\ y = positive\ \textbf{then}\)\\
\hskip8em   \(  C_{i+1} \leftarrow C_{i+1} \cup\{x\ channel\}\)\\
\hskip2em   \( \textbf{if}\ | C_{i+1} | > \tau\)\\
\hskip4em       \( C \leftarrow C \cup C_{i+1}\)\\
\hskip2em \textbf{else}\\
\hskip4em   stop process\\
\hskip2em   \(i \leftarrow i+1\)\\ 
 \caption{Channel Discovery Algorithm}
\label{algo1}
\end{algorithm}


\begin{algorithm}[h]
\SetAlgoLined
\(\textit{D} \leftarrow \emptyset \)\\
\(\textbf{for}\ x \in E\ \textbf{do}\)\\
\hskip2em  \(\textbf{if}\ x\ channel \in C\ \textbf{and}\ x\ channel \notin C_{1}\ \textbf{then}\) \\
\hskip4em  \(y \leftarrow chan2vec\_knn(L, E, x)\)\\
\hskip4em   \textbf{if} y = positive \textbf{then}\\
\hskip6em   \(\textit{D} \leftarrow D \cup \{x\quad channel \}\)\\
 \caption{Final Prediction}
 \label{algo2}
\end{algorithm}

\section{Experiments}

To show the effectiveness of our methods we run one channel discovery experiment and a variety of channel classification experiments. Since political content on YouTube has received much attention from researchers resulting in a variety of channel datasets, we decided to focus on this segment. However, we believe our method generalizes to a variety of other segments as well.

\subsection{Socio-Political Channel Discovery}
\label{recfluence}

Our first experiment involves discovering English language socio-political YouTube channels focused on the US. In order to do this we use the dataset created by Ledwich and Zaitsev (2019)\cite{ledwich2020algorithmic}. This dataset is continually being updated and we use a snapshot of it from July 27 2020. We will call this the Recfluence dataset after the research project data repository and use the channels 758 active channels in the set as positive instances for our political channel classification dataset.

Furthermore, we use a 773 channel dataset of non-political channels from Laukemper (2020) \cite{laukemper2020classifying} to get negative instances for the socio-political channel classification dataset. We call this dataset the Laukemper dataset after the author. Additionally, we also assume that all channels with over 3 million subscribers that are discovered and are not in the Recfluence set are negative examples i.e., not politically oriented content. This assumption is based on the fact that the Recfluence dataset was created in a way that ensures all of the most popular channels eligible for it were already discovered. 
We run our channel discovery process for four rounds with the following additional channel discovery details:

\begin{itemize}
    \item We use K=10 for all applications of KNN
    \item For each round, we use cross validation to measure the performance of chan2vec-knn on the labeled dataset using the latest embeddings. We choose a KNN threshold (the percentage of neighbors that must be positive to predict a channel is positive) that results in the highest possible precision while still having a recall >= 0.9. This will of course vary depending on the performance of the model and resources available for data collection.
    \item After a heuristically labeled negative channel has had commenter subscriptions queried, we add it to the labeled dataset.
\end{itemize}

After we finish identifying candidate channels, we make a final prediction for the candidate channels found during the four rounds. We use the 200 dimension embeddings generated during the final round of channel discovery, but also generate another set of 16 dimension embeddings. We apply KNN separately using each of these and average the results to generate a slight improvement in accuracy and a smoother precision / recall curve. We then filter out non-English channels using a method we will describe in the following section. Finally, we filter out channels with less than 20 commenter subscriptions since all channels in the Recfluence dataset had more than 20 commenter subscriptions.

Using a KNN threshold of 0.8, the final model has the following performance:
\begin{itemize}
    \item On the Laukemper + Recfluence Dataset
        \begin{itemize}
            \item Base Rate = 0.501
            \item AUC = 0.989
            \item Precision = 0.988
            \item Recall = 0.902
        \end{itemize}
    \item On the Laukemper + Recfluence + Heuristic Negative Channels Dataset
        \begin{itemize}
            \item Base Rate = 0.126
            \item AUC = 0.996
            \item Precision = 0.984
            \item Recall = 0.902
        \end{itemize}
\label{socio_political_class_results}
\end{itemize}

The results from each round of this process can be found in the Table \ref{tab:table}. "New Pos Channels" is the number of channels that were identified as candidate channels during the given round that the final model ultimately predicted to be positive. "Tot Subs for New Pos Channels" is the number of total subscribers for all of these channels. In the first round, these stats both only cover channels from the Recfluence dataset.

\begin{table}[H]
 \caption{Channel Discovery}
  \centering
  \begin{tabular}{p{1cm}|p{1.5cm}|p{1.5cm}|p{1cm}|p{1.5cm}}
   \toprule
    \scriptsize Channel Discovery Round   &   
    \scriptsize Total New Channels &   
    \scriptsize New Heuristic Negative Channels & 
    \scriptsize New Pos Channels & 
    \scriptsize Tot Subs for New Pos Channels\\
      \midrule
    1 & 1,449  & 0  & 758 & 431,550,776 \\
    2 & 16,153  & 3,643  & 5,259 & 489,726,779 \\
    3 & 8,965  & 790  & 1,732 & 115,027,796 \\
    4 & 3,079  & 25  & 233 & 33,284,599\\
    \bottomrule
  \end{tabular}
  \label{tab:table}
\end{table}

Despite only running for four rounds (including the initial round), this process discovers a very large number of channels:
\begin{itemize}
    \item 12.6M total channels
    \item 1.6M with the 5+ commenter subscriptions necessary for chan2vec
    \item 600K of these with 10K+ subscriptions (which is the threshold we set for the experiment)
    \item 25K of these identified by discovery process as candidate socio-political channels
    \item 7K of these are predicted to be socio-political by the final model
\end{itemize}

This immense number of channels shows why a machine learning approach like chan2vec-knn is necessary for classifying channels discovered through commenter subscriptions.

\subsection{Classifying Channel Language}

YouTube has over two billion monthly active users, and a significant amount of the content is not in English. The socio-political dataset we are using is limited to English language channels, and many of the heuristically added negative channels are non-English which leads to chan2vec-knn naturally classifying non-English channels as negative. However, many non-English channels are very political and are still close enough in the embedding space to be classified as socio-political.

Before we analyze the results of the socio-political channel discovery experiment further, we take the additional step of filtering out non-English channels. This is likely to be a necessary step for many channel discovery use cases and also presents an opportunity to show the effectiveness of chan2vec-knn on another task.

In order to generate a labeled dataset for classifying channel language we take advantage of the fact that the YouTube API provides information on default language\footnote{\href{https://developers.google.com/youtube/v3/docs/channels\#snippet.defaultLanguage}{https://developers.google.com/youtube/v3/docs/channels\#snippet.defaultLanguage}} for approximately five percent of channels. We gather default language data for 17K channels and label those with "en" and "en-GB" as English and all others as non-English. We then apply chan2vec-knn using the same embeddings generated in round four along with this new labeled dataset.

Using hold-one-out cross validation and a KNN threshold of 0.5 we find that this model has:
\begin{itemize}
    \item Base Rate = 0.630
    \item AUC = 0.937
    \item Precision = 0.859
    \item Recall = 0.946
\end{itemize}

We also check predictions for all channels in the Recfluence channel set, which is limited to positive examples, ie channels aimed at a English language audience. We find that only two out of 758 channels were predicted to be non-English and their predictions were both 0.5, which is right on our threshold. Both channels that the model predicted to be non-English were channels posting videos of foreign language communist anthems, so we find these classifications to be acceptable.

The Recfluence dataset only consists of positive examples, so it can be used to measure recall, but not precision, for the English language prediction model. The recall found on this dataset is much higher than it is on the training dataset, which we believe indicates the default language field used to create the training dataset might have a number of mislabeled channels. We save testing this hypothesis for future work.

As mentioned in the previous section, the model is applied to the set of 7,587 channels that the final classification model has predicted are socio-political. We find that 362 of these channels are predicted to be non-English and we remove them from the set. While this is only ~5\% of newly discovered channels, many of these had a large number of subscribers and would have impacted further analysis of socio-political YouTube, so it was worth the effort to remove them.

\subsection{Channel Discovery Performance Analysis}

The method used for discovering channels has two distinct parts where predictions are made.  The first part is a high recall model where the goal is to cast a wide net and discover as many candidate channels as possible, while still trying to avoid collecting too much unnecessary data. The second part is a high precision model where we use all the commenter subscription data we have collected to try to make as accurate of a prediction as possible about which candidate channels are positive (in the case of our experiment, which channels are socio-political).

To understand how well our method is performing, we focus on trying to measure the precision and recall. Since it is a two part method, the recall must be measured for each method. Since the second part is the only part that makes a final prediction, we only have to measure the precision on this.

\subsubsection{Out-of-sample Channel Analysis}

To better understand the performance of part two, our final socio-political classification model, we have three reviewers manually label a random sample of 100 of the 19,421 English candidate socio-political channels. After independent channel review and annotation, we reviewed the channel labels and discussed points of disagreement. During this review, we found that two of the channels no longer had videos available. For the remaining 98 channels, we can calculate the classification metrics. 

Our final socio-political classification model has:
\begin{itemize}
    \item Base Rate = 0.34
    \item Precision = 0.82
    \item Recall = 0.85
    \item ROC-AUC = 0.924
\end{itemize}

These metrics are lower than those found on the Recfluence and Laukemper dataset \ref{socio_political_class_results} 
, but that is expected given that a much smaller proportion of channels in that dataset were challenging borderline cases due to the way it was generated.

We also add the 100 new channels to our dataset and see how our model performs on them using hold-one-out cross validation:

\begin{itemize}
    \item Precision = 0.85
    \item Recall = 0.85
    \item ROC-AUC = 0.958
\end{itemize}

It is promising to see the increase in performance adding just 100 more labeled channels can produce.

\subsubsection{Recfluence Channel Hold Out Analysis}

The prior analysis helps us determine the precision and recall of the final socio-political classification model on candidate socio-political channels, but we still need to attempt to measure the recall of the first part of the method, the process that identifies candidate socio-political channels. Unfortunately there is no uniform sample of all YouTube channels we can label to determine this recall. Even if there was such a dataset, due to the low percentage of channels that are socio-political, we would not have the resources necessary to label enough channels to accurately determine this recall.

To get around this, we experiment with running the channel discovery process while holding out channels from the Recfluence set to see if they are discovered. Specifically, we use 5-fold cross-validation, where we remove 1/5 of the channels from the Recfluence set and see if the discovery process run using the other 4/5 of the Recfluence channels can discover the missing 1/5.

The computational resources and code complexity necessary to do this for all four rounds are high, so we limit ourselves to just the first round. We believe this is still informative given 70\% of the channels discovered were discovered in the first round.

We find that 92\% of held out Recfluence channels are found to be candidate socio-political channels in the first round. Given many channels are discovered in the subsequent rounds as well, we believe the recall would be even higher if we ran the experiment through all four rounds.

\subsubsection{Full Channel Discovery Precision and Recall}

We can then determine the final precision and recall numbers for the socio-political channel discovery process by combining the results from measuring the accuracy of both parts of the process. The precision is easy to determine since the second part of the process is the only part that can make a final positive prediction (the first part filters out channels, which is essentially making a negative prediction). Thus the precision of the overall process is the precision for the final socio-political model, which is 0.85 when including the new training data. The recall is the product of the recall of each step of the process, which is 0.92 * 0.85 = 0.78

Due to resource limitations, we are not able to get better measures of precision/recall for the full socio-political channels discovery/classification process than these. However, we believe these metrics point to strong enough performance by this model to use it for aggregate studies of the YouTube socio-political community.


\subsection{Classifying Recfluence Channel Tags}
\label{tag_classification}

In the previous experiment the goal was to discover channels that fit the criteria for Recfluence dataset of socio-political channels. We now turn to the task of classifying further information about these socio-political channels. We use the media type and soft tag labels specified for channels in the Recfluence dataset. In order to annotate the channels, each channel was reviewed by three manual reviewers. The study identified a set of 17 soft tags along with three media type tags. Each channel can only have one media type, but can have multiple soft tags. Many of the soft tags have a very limited number of examples in the Recfluence dataset, making it difficult to train and test models on them. Due to this, we limit ourselves to the 12 soft tags having greater than 30 instances in the dataset. Media Type and soft tags are briefly described in Table \ref{shorttags}. (For more detailed explanation of the tags, see Ledwich and Zaitsev (2019)). 

\begin{table}[h]
\caption{Media Type and Soft Tags}
\begin{tabular}{p{2.5cm}|p{5.5cm}}
\toprule
\textbf{Media Type} & \textbf{Description} \\
\midrule
\textbf{YouTube} &
Independent YouTube creators\\
\textbf{Mainstream Media} &
Content created by corporations such as cable news channels.\\
\textbf{Missing Link Media} & Channels funded by companies or venture capital, but not large enough to be considered “mainstream”. They are generally accepted as more credible than independent YouTube content. \\
\midrule
\textbf{Soft Tags} & \textbf{Description} \\
\midrule
\textbf{Partisan Left} &
Mainly focused on politics and exclusively critical of Republicans.\\
\textbf{Partisan Right} &
Mainly focused on politics and exclusively critical of Democrats and supports Trump.\\
\textbf{Conspiracy} & 
Regularly promotes a variety of conspiracy theories.\\
\textbf{Anti-SJW} & 
Have a significant focus on criticizing "Social Justice".\\
\textbf{Social Justice} & 
Espouses progressive views on social issues with a focus on identity politics, intersectionality, and political correctness. \\
\textbf{White Identitarian} &
Believes in and identifies with the superiority of "whites".\\
\textbf{Anti-theist} &
Self-identified atheist who are also actively critical of religion. \\
\textbf{Socialist} & 
Focused on problems with capitalism.\\
\textbf{Religious Conservative} &
Focused on promoting Christianity or Judaism in the context of politics / culture. \\
\textbf{Libertarian} &
Focused on individual liberties and generally skeptical of authority and state power. \\
\textbf{Educational} &
Has significant focus on educational material related to politics / culture. \\
\textbf{State Funded} &
Channels funded by a government. \\
\bottomrule
\end{tabular}
\label{shorttags}
\end{table}

To predict soft tags and media types of channels we use chan2vec-knn with the embeddings generated from the last round of the channel discovery experiment and the Recfluence tag labels as our labeled dataset. Previously we have used K=10, but due to the small number of channels with some labels, we use K=5 for this experiment. Metrics for this experiment can be found in Table \ref{tab:recallprecisionsmall}.

\begin{table}[h]
 \caption{Tag Classification Performance}
  \centering
  \begin{tabular}{p{2.5cm}|p{0.75cm}|p{0.75cm}|p{0.75cm}|p{0.75cm}|p{0.75cm}}
   \toprule
    \scriptsize Tag   &   
    \scriptsize \# Channels with Label &   
    \scriptsize Precision & 
    \scriptsize Recall &
    \scriptsize Reviewer Agreement & 
    \scriptsize Model Agreement \\
      \midrule
    YouTube & 646 & 0.95 & 0.96 & - & - \\
    Mainstream Media & 80 & 0.69 & 0.76 & - & - \\
    Missing Link Media & 36 & 0.33 & 0.06 & \textbf{95.4\%} & 94.6\% \\
    \midrule
    Partisan Right & 209 & 0.78 & 0.83 & 81.2\% & \textbf{84.0\%} \\
    Partisan Left & 124 & 0.71 & 0.68 & 85.6\% & \textbf{87.2\%} \\
    Social Justice & 107 & 0.71 & 0.59 & 87.8\% & \textbf{89.3\%} \\
    AntiSJW & 243 & 0.78 & 0.83 & 83.0\% & \textbf{85.0\%} \\
    Conspiracy & 78 & 0.89 & 0.82 & 95.3\% & \textbf{95.9\%} \\
    Religious Conser.. & 47 & 0.67 & 0.17 & 92.7\% & \textbf{93.4\%} \\
    Anti-Theist & 44 & 0.68 & 0.89 & 97.0\% & \textbf{97.5\%} \\
    Socialist & 44 & 0.90 & 0.80 & 94.2\% & \textbf{95.1\%} \\
    Libertarian & 37 & 0.71 & 0.41 & 93.7\% & \textbf{94.3\%} \\
    Educational & 33 & 0.92 & 0.33 & 91.0\% & \textbf{93.9\%} \\
    White Identitarian & 32 & 0.69 & 0.69 & 96.8\% & \textbf{97.0\%} \\
    State Funded & 31 & 0.71 & 0.32 & \textbf{99.1\%} & 96.7\% \\
    \bottomrule
  \end{tabular}
  \label{tab:recallprecisionsmall}
\end{table}

We look at the precision and recall of each tag classifier as well as metrics around tag agreement. This includes "Reviewer Agreement", which is the percentage of time all pairs of manual reviewers agreed with each other on whether a channel should have a given tag as well as "Model Agreement" which is the percentage of time predictions by the model agreed with each manual reviewer for each channel. 
Since YouTube and Mainstream Media tags were annotated using a source other than manual reviewers, they do not have agreement metrics. We find that for the 13 tags that have agreement metrics, the model has a higher agreement percentage than the average individual reviewer for 11 of them. This is in part due to the impressive performance of the model, but the difficulty of manual annotation for many of these tags leading to low reviewer agreement contributes to this as well. The poor performance by the model on the "Missing Link Media" and "State Funded" tags is also understandable. These tags are meant to capture the production value and funding source of a channel. Neither of these factors are likely to lead to channels with the these tags having similar subscribers. For example, BBC News and RT are both "State Funded", but likely have very different sets of subscribers.

\subsection{Analyzing Newly Discovered Socio-political Channels}

We now combine the results from socio-political channel discovery and tag classification to show an example of the benefit of using these methods for more comprehensive examinations of communities on YouTube. Specifically, we attempt to determine the size of communities represented by the Recfluence soft tags and show that limiting an analysis of these communities to just the most popular socio-political channels creates an inaccurate picture of how these communities compare in size.

We first apply the chan2vec-knn tag classification approach from the previous section to the 7,225 newly discovered socio-political channels in order to get tag predictions for them. We then combine the new socio-political channels with the Recfluence channels and analyze the amount of traffic for all channels (new and from Recfluence) with a given soft tag. Since many soft tags have different false positive and false negative rates, we also account for this using a method described in Appendix \ref{multiplier} to determine a multiplier used to adjust soft tag group estimates. For traffic, we use SocialBlade\footnote{\href{https://socialblade.com/}{https://socialblade.com/}} to get the number views for each channel over the previous 12 months.

\begin{figure}[ht]
\centering
    \includegraphics[width=\linewidth]{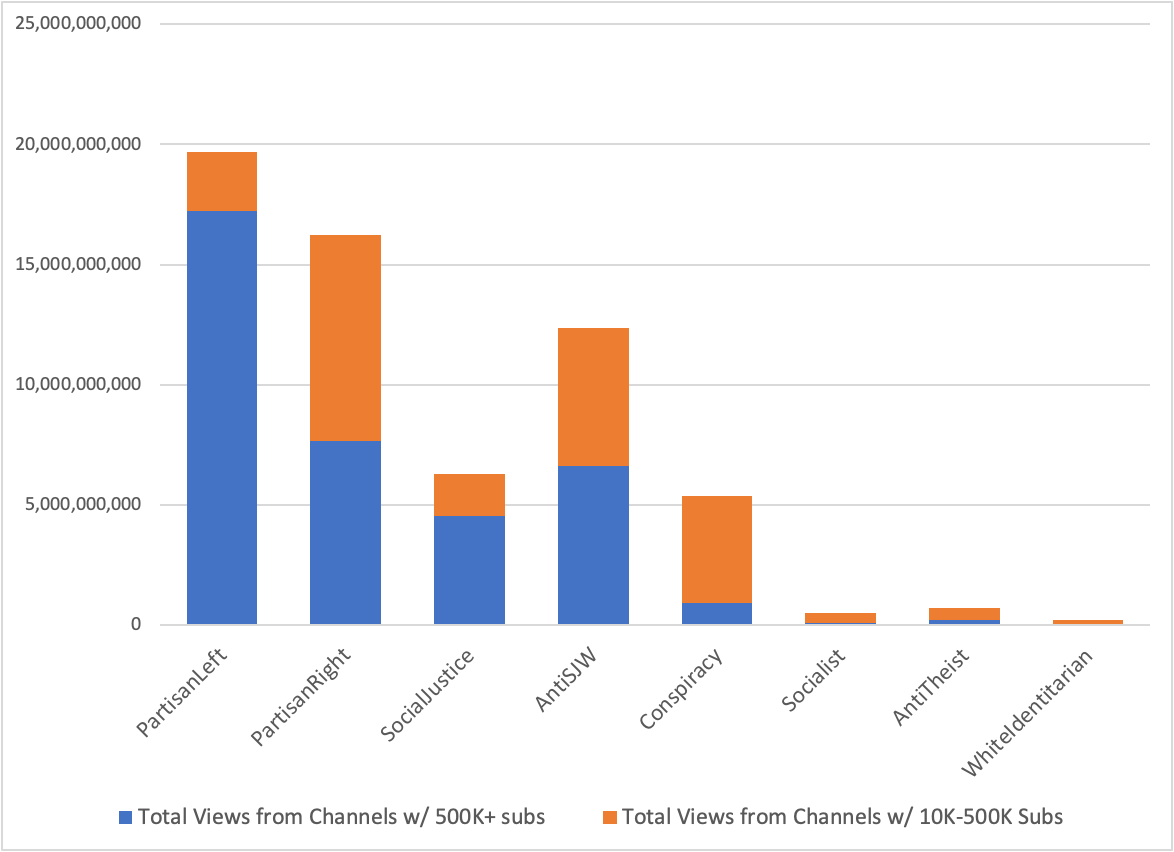}
    \caption{Head vs. Tail Channel Aggregate Views by Soft Tag}
    \label{HeadVsTail}
\end{figure}

One common method for analyzing YouTube is to focus on the most popular channels, also referred to as head channels. This was done explicitly in a recent Pew study \cite{stocking2020share} and indirectly by Ledwich and Zaitsev (2019) through their method of channel discovery. While Ledwich and Zaitsev (2019) used their dataset to measure bias in the recommendation system, it is reasonable to see how one might attempt to use it to compare the size of socio-political groups on YouTube. We next try to determine what coverage can be obtained by these head channels and how representative they are of all socio-political YouTube. We do this by comparing the 237 channels with over 500,000 subscribers to the 8,044 channels with between 10,000 and 500,000 subscribers in Figure \ref{HeadVsTail} and Appendix \ref{tab:head_vs_tail}.

We find that while head channels account for over 60\% of the overall socio-political traffic, using head channels alone to compare the size of socio-political communities results in significantly different conclusions than analyzing a more comprehensive set of socio-political channels. Specifically, for head channels alone, it appears that ’Partisan Left’ has over twice the traffic of ’Partisan Right’ and ’Conspiracy’ channels is a very distant fifth. While looking at views from all socio-political channels shows that ’Partisan Left’ only has 30\% more aggregate views than ’Partisan Right’ and ’Conspiracy’ has only 15\% fewer aggregate views than ’Social Justice’.


\subsection{Classifying Political Leaning from MBFC Dataset}
\label{dinkovcomparsion}

In our final experiment we compare the performance of chan2vec-knn to another channel classification approach introduced by Dinkov et al's (2019) \cite{dinkov2019predicting} that uses video metadata, transcripts, and audio as input data. We use a dataset containing the political lean of news channels. Dinkov et al. created this dataset by connecting media outlets that have had their political leanings annotated by Media Bias/Fact Check (MBFC) website \footnote{\href{https://mediabiasfactcheck.com/}{https://mediabiasfactcheck.com/}}, an online resource for mainstream media classification, to their corresponding YouTube channel. This resulted in a dataset with 421 channels in which each channel is given a left, center, or the right label.

We compare our results to their best performing model and find this to be a valuable benchmark given the state of the art NLP approach used to create this model and the impressive results they obtained from including audio signals.

First, we eliminate the nine channels from the dataset which have been deactivated since it was created. We are left with 412 channels for method comparison. Since the chan2vec-knn approach requires a minimum number of commenter subscriptions to be collected for a channel in order to classify that channel and the MBFC dataset contains many channels with under 1,000 subscribers, we are not able to make predictions for all 412 channels.  In order to properly compare our methods, we reproduce Dinkov et al.'s (2019) predictions and then only measure their performance on channels chan2vec-knn can make predictions for.

There are some ways we can increase commenter subscription data beyond what is described in the prior data collection section.  We experiment with four different data collection methods and share the results for how chan2vec-knn and the Dinkov et al. model performs on each of the resulting channel sets, along with showing the performance of their model on the full 412 channel set.  These metrics are shared in Table \ref{tab:table1}.

The commenter subscription data collection methods we use are:
\begin{enumerate}
    \item We follow the exact method described in the data collection section \ref{datacollection} and gather commenter subscriptions for all 412 channels.
    \item We use all data from (1) along with gathering full commenter subscriptions for all commenters instead of just the ten we normally sample.
    \item We use all data from (2) along with gathering data using a standard data collection method for the set of channels in the Recfluence dataset.
    \item We use all data from (3) along with gathering data on 10,561 channels identified as socio-political in the channel discovery experiment.
\end{enumerate}

\begin{table*}[ht]
 \caption{MBFC Political Lean Prediction Accuracy Comparison}
  \centering
  \begin{tabular}{p{3cm}|p{1.5cm}|p{1.5cm}|p{1.5cm}|p{1.5cm}|p{1.5cm}|p{1.5cm}|p{2cm}}
    \toprule
    \multicolumn{2}{c}{Part}                   \\
  Commenter Subscription Set   & 
  Number of Channels Supported & Number of Channels with 1K+ Subs &
  Number of Channels w/ 10K+ Subs & 
  chan2vec - multi-class accuracy & chan2vec - regression accuracy & 
  Dinkov et al. accuracy & 
  Total Subs Supported\\
    \midrule
    1) MBFC Channel Normal Subs & 202  & 183  & 124 & \textbf{83.2} & 81.7 & 72.8 & 45,733,303\\
    2) (1) + All Full Subs & 261  & 227  & 138 & \textbf{85.8} & 85.4 &  70.9 & 46,822,049\\
    3) (2) + All Socio-political Channel Subs & 278  & 237  & 140 & 84.2 & \textbf{85.3} & 70.5 & 46,879,831\\
    4) (3) + All New Political Channel Subs & 333  & 265  & 142 & 81.7 & \textbf{83.8} & 73.0 & 47,003,925\\
    \midrule
    Dinkov et al - Full MBFC Predictions & 412 & 270 & 142 & - & - & \textbf{73.3} & 47,012,187\\
   \bottomrule
  \end{tabular}
  \label{tab:table1}
\end{table*}

We then report the following information about each data collection method
\begin{itemize}
    \item "Number of Channels Supported" is the number of channels from the MBFC set that the data collection method was able to gather enough commenter subscriptions for to enable chan2vec-knn to make predictions.
    \item "Number of Channels with 1K+ and 10K+ subs" are columns that share the number of channels with over 1,000 subscribers or over 10,000 subscribers that chan2vec-knn was able to make predictions for.
    \item "Chan2vec multi-class accuracy" is the accuracy of the model when treating the left/center/right prediction task as a multi-class classification task, meaning the model is not aware that they are points along a spectrum.  The prediction is the label most common in the ten nearest neighbors.
    \item "Chan2vec regression accuracy" is the accuracy of the model when treating the left/center/right prediction task as a regression task. In this case, the left is -1, the center is 0, and the right is 1.  The prediction is the weighted average of the ten nearest neighbors rounded to the nearest integer.
    \item "Dinkov et al. accuracy" is the accuracy of the Dinkov model predictions on the set channels the chan2vec model made predictions on.
    \item "Total Subs Supported" is the total number of subscribers for all channels supported in the given experiment. 
\end{itemize}

From the Table \ref{tab:table1}, we can see that both implementations of chan2vec-knn significantly outperform the Dinkov model across all channel subsets predictions were made on.

The Dinkov model does have the ability to make predictions for many small channels that the chan2vec-knn approach can not, but the chan2vec-knn approach has good coverage for channels with over 10,000 subscribers.

Specifically: 
\begin{itemize}
    \item model (2), which only gathers commenter subscription information from MBFC channels, can cover 97\% of channels with over 10,000 subscribers and 84\% of channels with over 1,000 subscribers.  Even better, 99.60\% of total subscribers are covered by these predictions. 
    \item model (4), which uses significantly more commenter subscription data, can cover 100\% of channels with over 10,000 subscribers and 98\% of channels with over 1,000 subscribers, and 99.98\% of total subscribers.
\end{itemize}

It is also interesting to note how the coverage of the MBFC channels grows as commenter subscriptions from new channels are added. The socio-political channels included were not perfect matches for the MBFC dataset, but there is enough overlap in subscribers that it resulted in a 19\% increase in supported MBFC channels.

\section{Conclusion}

This study makes a variety of contributions to the YouTube research space. We present new methods for automatically classifying and discovering YouTube channels to help with the challenge of analyzing communities and categories at the platform's scale. Due to the abundance of political YouTube datasets, we focus on this category to test our methods, but we believe these methods can be applied to a variety of other categories.

Multiple experiments are run to test the performance of chan2vec-knn channel classification. Using the Recfluence dataset \cite{ledwich2020algorithmic}, we find that chan2vec-knn is able to predict "soft tags" for channels more accurately than the average individual human annotator for 11 / 12 tags in the experiment. Additionally, seven of these tags have less than fifty examples in the dataset, which demonstrates that the method can perform quite well with a limited amount of labeled data. We are also able to compare chan2vec-knn to the Dinkov et al machine learning approach to channel classification. Chan2vec-knn shows a sizeable improvement over their model, achieving an accuracy of 83.8\% (compared to 73.0\%) on the task of predicting the political lean of news channels. Since chan2vec-knn relies on commenter subscriptions, it is not able to make predictions for 79 of the 412 channels in the Dinkov et al. dataset, however it is able to make predictions for 265 out of the 270 channels that have over 1,000 subscribers. Furthermore, the channels supported by chan2vec-knn cover 99.98\% of the total subscribers for channels in the dataset.

A single experiment to test the performance of our channel discovery method is run. We use the method to identify socio-political channels that match the criteria for the Recfluence dataset. The method discovers 7,224 such channels, which is nearly 10x the 758 channels in the original dataset. Unlike all other studies we have encountered, we go a step further and attempt to measure the percentage of all socio-political channels that our new set of channels, in addition to the original dataset, covers. Using an analysis where we hold out channels from the original dataset when running the discovery method, we estimate that our new dataset covers 78\% of channels that match the criteria. Although, due to limitations of this analysis, we believe the coverage might be even higher.

These newly discovered channels are then used in a short analysis to show the importance of doing comprehensive studies of YouTube categories instead of just looking at the most popular channels in a category. We do this by classifying "soft tags" for the newly discovered channels then combining these with the original Recfluence dataset to create a full set of socio-political channels. The 257 channels with over 500,000 subscribers, which we refer to as "head" channels, are then compared to the entire set of channels. We find that only 1.8\% of traffic to "head" channels goes to "Conspiracy" channels, while 6.5\% of overall socio-political traffic goes to "Conspiracy" channels. Similarly, 14.6\% of "head" traffic goes to "Partisan Right" channels, while this rises to 19.8\% when considering overall traffic. This means that a study that only analyzed the "head" socio-political channels would significantly underestimate the amount of traffic going to "Conspiracy" and "Partisan Right" channels.

\section{Limitations and Future Work}

There are some limitations to our study that must be considered. First, like many other YouTube researchers, we analyze content at the channel level instead of the video level. This approach functions for the majority of content, which comes from channels with narrow themes. However, there are some channels with a large breadth of content that are not handled well by this approach. Second, our method relies entirely on subscribers to determine embeddings and embeddings to classify the type of content a channel has.  The experiments we have run have shown that embeddings near each other tend to have the same content, however it is possible there are many tasks where two different but highly correlated types of content would easily be misclassified. Finally, our method can only be applied to channels with above a certain number of subscribers. More experimentation needs to be done to determine how low this threshold is, but it is very unlikely channels with less than 100 subscribers will be discovered or classified correctly.

For the brevity of this study, we limited our approach to commenter subscriptions, but we believe there is certainly room for improvement by leveraging other data. Given how orthogonal the approaches taken by Dinkov et al. (2019) and others are to chan2vec, we believe an improvement in performance can easily be achieved from ensembling our methods. 

\subsection{Code Access}

We open-sourced our code for the experiments described in this paper, and it can be found on Github\footnote{\href{https://github.com/sam-clark/chan2vec}{https://github.com/sam-clark/chan2vec}}. 







\subsection{Acknowledgments}

We would like to thank Mark Ledwich and Anton Laukemper for sharing data, help with annotation, and helpful discussions.

\bibliographystyle{unsrt}  
\bibliography{./bibliography/youtube}  



%
%
%

\vspace{5mm}
\begin{appendices}





\section{Soft tag Precision and Recall Normalization}
\label{multiplier}
\noindent The recall from the hold-out analysis of the channel discovery process is presented by each tag in table \ref{tab:recfluence}.

\begin{table}[H]
 \caption{Soft Tag Recall Comparison}
  \centering
  \begin{tabular}{p{3cm}|p{1.3cm}|p{1.3cm}|p{1.3cm}}
   \toprule
    \scriptsize Tag   &   
    \scriptsize \# Channels with Tags &   
    \scriptsize \# Chans - Candidate Found & 
    \scriptsize \% Chans - Candidate Found \\ 
      \midrule
    YouTube & 645 & 604 & 94\% \\
    Mainstream Media & 79 & 65 & 82\%\\
    Missing Link Media& 35 & 27 & 75\%\\
    \midrule
    AntiSWJ & 243 & 237 & 98\%\\
    Partisan Right & 209 & 205 &98\%\\
    Partisan Left & 122 & 110 & 90\%\\
    Social Justice & 106 & 94 & 89\%\\
    Conspiracy & 78 & 74 & 95\%\\
    Religious Conservative & 47 & 43 & 92\%\\
    Anti-Theist & 44 & 44 & 100\%\\
    Socialist & 44 & 42 & 96\%\\
    Libertarian & 37 & 36 & 97\%\\
    Educational & 33 & 27 & 82\%\\
    White Identitarian & 32 & 32 & 100\%\\
    State Funded & 31 & 27 & 87\%\\
    \bottomrule
  \end{tabular}
  \label{tab:recfluence}
\end{table}

To accurately estimate group sizes, we must account for the precision and recall of our models and how this impacts the false negative and false positive rates. For a given tag we get the following:
\begin{itemize}
    \item Recall = Political Channel Classification Recall * Soft Tag Classification Recall
    \item Precision = Political Channel Classification Precision * Soft Tag Classification Precision
    \item Multiplier = Precision / Recall
\end{itemize}

\section{Head versus Tail Socio-Political Channels}

\noindent Table \ref{tab:head_vs_tail} presents the values for multipliers for each tag along with other tag statistics. The method assumes that the political classification and soft tag predictions are independent, which is unlikely to be completely true. However, without this assumption, it would be too costly to label the necessary data to measure the true precision and recall of the final soft tag predictions, so we must settle for it. The multiplier for a soft tag is the number we multiply by the aggregate number of new channels, subscribers, and views in order to account for the difference in false positive and false negative rates for the entire channel discovery and soft tag prediction process.

\clearpage
\begin{table*}[t]
 \caption{Channels 500K+ Subs vs. Channels 10K-500K Subs - Last 12 month Views }
  \centering
  \begin{tabular}{p{2cm}|p{1cm}|p{1cm}|p{2cm}|p{1cm}|p{2cm}|p{1cm}|p{2cm}|p{1cm}}
   \toprule
Tag & Multiplier & \% all Political Views & Total Tag Views & Channels w/ 500K+ Subs & Total Views from Channels w/ 500K+ subs &  Channels w/ 10K-500K Subs & Total Views from Channels w/ 10K-500K Subs & \% Overall Views from Channels w/ 500K+ Subs \\
\midrule
YouTube & 1.06 & 46.1\% & 38,409,907,772 & 152 & 19,024,388,129 & 7,248 & 19,385,519,642 & 50\%	\\
\midrule
MainstreamMedia & 1.11 & 52.2\% & 43,495,914,582 & 85 & 33,264,111,364 & 796 & 10,231,803,217 & 76\% \\
\midrule
PartisanLeft & 1.17 & 23.6\% & 19,699,048,250 & 48 & 17,218,750,066 & 588 & 2,480,298,183 & 87\% \\
\midrule
PartisanRight & 0.96 & 19.5\% & 16,215,133,594 & 45 & 7,649,620,820 & 2,569 & 8,565,512,773 & 47\% \\
\midrule
SocialJustice & 1.36 & 7.5\% & 6,271,926,124 & 31 & 4,544,912,590 & 703 & 1,727,013,533 & 72\% \\
\midrule
AntiSJW & 0.97 & 14.8\% & 12,346,433,837 & 57 & 6,630,044,828 & 1,434 & 5,716,389,009 & 54\% \\
\midrule
Conspiracy & 1.14 & 6.4\% & 5,360,279,151 & 26 & 930,442,686 & 2,744 & 4,429,836,465 & 17\% \\
\midrule
Socialist & 0.81 & 0.6\% & 500,209,308 & 3 & 99,764,697 & 158 & 400,444,611 & 20\% \\
\midrule
AntiTheist & 1.13 & 0.9\% & 718,463,993 & 5 & 232,762,040 & 271 & 485,701,953 & 32\% \\
\midrule
WhiteIdentitarian & 1.00 & 0.3\% & 229,883,175 & 2 & 6,374,859 & 124 & 223,508,316 & 3\% \\
\bottomrule
  \end{tabular}
  \label{tab:head_vs_tail}
\end{table*}

\end{appendices}
\end{document}